# Contracting and Involutive Negations of Probability Distributions

Ildar Batyrshin

Instituto Politécnico Nacional, Centro de Investigación en Computación, CDMX, México
batyr1@gmail.com

**Abstract.** A dozen papers have considered the concept of negation of probability distributions (pd) introduced by Yager. Usually, such negations are generated point-by-point by functions defined on a set of probability values and called here negators. Recently it was shown that Yager's negator plays a crucial role in the definition of pd-independent linear negators: any linear negator is a function of Yager's negator. Here, we prove that the sequence of multiple negations of pd generated by a linear negator converges to the uniform distribution with maximal entropy. We show that any pd-independent negator is non-involutive, and any non-trivial linear negator is strictly contracting. Finally, we introduce an involutive negator in the class of pd-dependent negators that generates an involutive negation of probability distributions.

## 1. Introduction

The concept of the negation of a probability distribution (pd) was introduced by Yager [1]. He defined the negation $\bar{P}$ of a finite probability distribution $P = (p_1, ..., p_n)$ by: $\bar{P} = (\bar{p_1}, ..., \bar{p_n})$, where $\bar{p_i}$ is defined by: $\bar{p_i} = \frac{1-p_i}{n-1}$. He noted that other negations of probability distributions (pd) could exist. Further, negations of probability distributions were considered in many works [2-12]. The properties of Yager's negation of a probability distribution are studied in [2]. The paper [3] studied uncertainty related to Yager's negation. The authors of [4, 5] studied the convergence of the sequence of multiple Yager's negations of pd to the uniform distribution. In [6], Yager's negation is used in a multi-criteria decision-making procedure. In [7], the authors introduced another negation of probability distributions based on Tsallis entropy. The authors of [8] considered a negation of basic probability assignment in Dempster-Shafer theory. The authors of [9] studied the properties of the negation of basic probability assignment based on total uncertainty measure. The paper [10] gives a definition of negation in basic belief assignment in the Dempster–Shafer theory using matrix operators. This matrix negation was also considered in [11]. The authors of the paper [12] studied functions called negators defined on the set of probability values and point-by-point transforming pd in its negation. The results of this paper will be used here. Two types of negators are considered: pd-independent and pd-dependent, and a class of pd-independent linear negators was introduced [12]. It was shown [12] that Yager's negator plays a crucial role in the definition of pd-independent linear negators: any linear negator is a function of Yager's negator.

In this paper, we study contracting and involutive negators of pd. Involutive negations in fuzzy logic were considered in [13]. Contracting negations were introduced in [14] for lexicographic valuations of plausibility values used in various expert systems with qualitative expert opinions [15]. Further such negations have been studied in fuzzy logic [16-18]. In this paper, we prove that the sequence of multiple negations of pd generated by a linear negator converges to the uniform distribution with maximal entropy. We show that any pd-independent negator is non-involutive, and any non-trivial linear negator is strictly contracting. Finally, we introduce an involutive negator in the class of pd-dependent negators that generates an involutive negation of probability distributions.

The paper has the following structure. Section 2 considers basic definitions of negators and negations of pd generated by negators. Section 3 describes pd-independent and linear negators' properties from [12] used in the following sections. Section 4 introduces a general form of multiple linear negators and finds the limit of the sequence of such multiple negators. Section 5 shows non-involutivity of pd-independent negators, and shows that non-trivial linear negators are strictly contracting. In Section 6, we introduce an involutive pd-dependent negator that defines the involutive negation of probability distribution. The last section contains conclusions.

## 2. Negations of Discrete Probability Distributions

A set $P = \{p_1, \ldots, p_n\}$, $(n \geq 2)$, of $n$ real values $p_i$ satisfying for all $i = 1, \ldots, n$, the properties:

$$0 \leq p_i \leq 1, \quad \sum_{i=1}^{n} p_i = 1, \tag{1}$$

is referred to as a (finite discrete) *probability distribution* (*pd*) *of the length n*. One can consider $p_i$ as a probability of an outcome $x_i$ in some experiment $X$ with outcomes $\{x_1, \ldots, x_n\}, n \geq 2$. Let $\mathcal{P}_n$ be the set of all possible probability distributions of the length $n$ defined on $X$. For simplicity of the interpretation; we will fix the ordering of outcomes and the ordering of corresponding probability values according to their indexing $i = 1, \ldots, n$, and represent the probability distribution $P = \{p_1, \ldots, p_n\}$ as $n$-tuple $P = (p_1, \ldots, p_n)$.

The probability distribution $P_{(i)} = (p_1, \ldots, p_n)$ satisfying the property: $p_i = 1$ for some $i$ in $\{1, \ldots, n\}$, and $p_j = 0$ for all $j \neq i$, will be referred to as a *degenerate* or *point distribution* [19]. For example, for $i = 1$ and $i = n$ we have the following point distributions: $P_{(1)} = (1,0,\ldots,0)$, and $P_{(n)} = (0,\ldots,0,1)$.

The simplest example of pd is the *uniform distribution*: $P_U = \left(\frac{1}{n}, \ldots, \frac{1}{n}\right)$.

A transformation $NOT(P)$ of probability distributions $P = (p_1, \ldots, p_n)$ from $\mathcal{P}_n$ into probability distributions $NOT(P) = Q = (q_1, \ldots, q_n)$ in $\mathcal{P}_n$ is called a *negation of probability distributions* if, for all $i, j = 1, \ldots, n$, the following properties are satisfied:

$$0 \leq q_i \leq 1, \quad \sum_{i=1}^{n} q_i = 1, \tag{2}$$

$$\text{if } p_i \leq p_j, \text{ then } q_i \geq q_j, \tag{3}$$

From (3) it follows for all $i, j = 1, \ldots, n$:

$$\text{if } p_i = p_j, \text{ then } q_i = q_j. \tag{4}$$

A *negator* $N$ is a function of probability values $p_i$ point-by-point transforming probability distributions $P = (p_1, \ldots, p_n)$ into probability distributions: $NOT(P) = (N(p_1), \ldots, N(p_n))$, hence, for all $i = 1, \ldots, n$, the following properties are satisfied:

$$0 \leq N(p_i) \leq 1, \quad \sum_{i=1}^{n} N(p_i) = 1, \tag{5}$$

$$\text{if } p_i \leq p_j, \text{ then } N(p_i) \geq N(p_j). \tag{6}$$

We will say that a negator $N$ *generates* (point-by-point) a negation $NOT(P) = (N(p_1), \ldots, N(p_n))$ of pd $P$.

A negator $N$ is called a *pd-independent* [12] if for any pd $P = (p_1, \ldots, p_n)$ in $\mathcal{P}_n$ the negator $N(p_i)$ depends only on the value $p_i$ but not on other values $p_j$ from $P$. A negator that is not pd-independent will be referred to as *pd-dependent*. A negation $NOT_N(P) = (N(p_1), ..., N(p_n))$ of a probability distribution $P = (p_1, ..., p_n)$ will be called a *pd-independent negation* of pd if it is generated by pd-independent negator $N$.

Yager's negator [1]:

$$N_Y(p) = \frac{1-p}{n-1}, \text{ for all p in } [0,1], \tag{7}$$

is a pd-independent negator. For any pd $P = (p_1, ..., p_n)$ in $\mathcal{P}_n$ it defines negation of $P$:

$$NOT_Y(P) = (N_Y(p_1), ..., N_Y(p_n)) = \left(\frac{1-p_1}{n-1}, ..., \frac{1-p_n}{n-1}\right).$$

The *uniform negator* [12]:

$$N_U(p) = \frac{1}{n}, \text{ for all } p \text{ in } [0,1], \tag{8}$$

is another example of pd-independent negator. For any pd $P = (p_1, ..., p_n)$ in $\mathcal{P}_n$ negator $N_U$ defines its negation: $NOT_U(P) = (N_U(p_1), ..., N_U(p_n)) = \left(\frac{1}{n}, ..., \frac{1}{n}\right) = P_U$, where $P_U = \left(\frac{1}{n}, ..., \frac{1}{n}\right)$ is the *uniform distribution*.

The following negator, introduced by Zhang et all. [7]:

$$N_T(p_i) = \frac{1-p_i^k}{n - \sum_{j=1}^n p_j^k}, k \neq 0.$$

is an example of pd-dependent negator.

In the following sections, we will show that all pd-independent negators are non-involutive. We will introduce an involutive negator in the class of pd-dependent negators. This involutive negator will generate an involutive negation of probability distributions satisfying the property: $NOT(NOT(P)) = P$.

### 3. Properties of Pd-Independent and Linear Negators

In [12] it was shown that Yager's negator plays a crucial role in the construction of pd-independent linear negators: any linear negator is a function of Yager's negator. Let us consider some properties of pd-independent and linear negators that will be used further in this paper.

An element $p$ in $[0,1]$ is called a *fixed point* of a negator $N$ if $N(p) = p$. A probability distribution $P$ is called a *fixed point* of a negation $NOT$ if $NOT(P) = P$.

**Proposition 1** [12]. *Any negator $N$ has a fixed point $p = \frac{1}{n}$, i.e., $N\left(\frac{1}{n}\right) = \frac{1}{n}$, and the negation of probability distribution $NOT(P) = (N(p_1), ..., N(p_n))$ generated by $N$ have the fixed point:*

$$NOT(P_U) = P_U. \tag{9}$$

**Theorem 1** [12]. *Any pd-independent negator $N$ has the unique fixed point $p = \frac{1}{n}$, and any pd-independent negation of probability distributions $NOT_N$ has a unique fixed point $P_U$.*

**Corollary 1.** *Any pd-independent negator $N$ satisfies the following property:*

$$N(x) = x, \text{ if and only if, } x = \frac{1}{n}. \tag{10}$$

**Proposition 2** [12]. *Any pd-independent negator N satisfies the property:*

$$N(0) = \frac{1-N(1)}{n-1}. \tag{11}$$

From (11), we obtain:

$$N(1) = 1 - (n-1)N(0). \tag{12}$$

**Proposition 3** [12]. *Any pd-independent negator N satisfies the following properties:*

$$N(1) \in \left[0, \frac{1}{n}\right], \tag{13}$$

$$N(0) \in \left[\frac{1}{n}, \frac{1}{n-1}\right]. \tag{14}$$

**Theorem 2** [12]. *Any pd-independent negator N satisfies the following properties:*

$$N(p) \in [0, \frac{1}{n}] \quad \text{if } p \geq \frac{1}{n}, \tag{15}$$

$$N(p) \in [\frac{1}{n}, \frac{1}{n-1}] \quad \text{if } p \leq \frac{1}{n}. \tag{16}$$

**Definition 1** [12]. A pd-independent negator $N$ is referred to as a *linear negator* if $N(p)$ is a linear function of $p \in [0,1]$. The negation $NOT_N(P) = (N(p_1), \ldots, N(p_n))$ of a probability distribution $P = (p_1, \ldots, p_n)$ is called a *linear negation* of pd if $N(p)$ is a linear negator.

**Theorem 3** [12]. *A function $N(p)$ is a linear negator if and only if it is a convex combination of negators $N_U$ and $N_Y$, i.e., for some $\alpha \in [0,1]$ for all p in [0,1] the following property is satisfied:*

$$N(p) = \alpha N_U(p) + (1-\alpha) N_Y(p) = \alpha \frac{1}{n} + (1-\alpha) \frac{1-p}{n-1}, \tag{17}$$

*where $\alpha \in [0,1]$ is a parameter of the convex combination.*

From (13), we have $nN(1) \in [0,1]$, hence in (17), we can use $\alpha = nN(1)$ and represent (17) as a function of Yager's negator $N_Y$ [12]:

$$N(p) = N(1) + (1 - nN(1))\frac{1-p}{n-1} = N(1) + (1 - nN(1))N_Y(p), \tag{18}$$

where $N(1)$ is a parameter $N(1) \in \left[0, \frac{1}{n}\right]$, defining the value of the negator $N(1)$ for $p = 1$.

The paper [12] formulates an Open Problem: Prove or disprove a hypothesis that any pd-independent negator is linear. We think that any pd-independent negator is linear. In such a case, the properties established in [12] and in this paper for pd-independent negators will be fulfilled for linear negators and vise versa.

The following section shows that the sequence of multiple linear negations of a pd converges to the uniform distribution with the maximal entropy.

4. **Multiple Linear Negators and Negations**

For all $k = 1,2,\ldots$, denote $N^k(p) = N\left(N^{k-1}(p)\right)$, where $N^0(p) = p$ and $N^1(p) = N(p)$. We have:

$$N^{k+2}(p) = N\left(N^{k+1}(p)\right) = N\left(N\left(N^k(p)\right)\right) = N^2\left(N^k(p)\right).$$

For any $p$ in [0,1] denote $d = p - \frac{1}{n}$, then $p = \frac{1}{n} + d$.

Applying (12): $N(1) = 1 - (n-1)N(0)$, in linear negator (18): $N(p) = N(1) + (1 - nN(1))\frac{1-p}{n-1}$, after equivalent transformations, we obtain another representation of linear negator:

$$N(p) = N(0) + (1 - nN(0))p. \tag{19}$$

Substituting in (19): $p = \frac{1}{n} + d$ and $A = 1 - nN(0)$ we obtain:

$$N(p) = N(0) + (1 - nN(0))\left(\frac{1}{n} + d\right) = N(0) + \frac{1}{n} - nN(0)\frac{1}{n} + Ad = \frac{1}{n} + Ad. \tag{20}$$

**Proposition 4.** *For linear negators $N$, the following formulas hold for any $p$ in [0,1] and for all $k = 0, 1, 2, \ldots$:*

$$N^k(p) = \frac{1}{n} + A^k d, \quad \text{where } A = 1 - nN(0), \quad \text{and} \quad d = p - \frac{1}{n}. \tag{21}$$

**Proof.** (21) holds for $k = 0$: $N^0(p) = p = \frac{1}{n} + A^0 d = \frac{1}{n} + p - \frac{1}{n} = p$. (21) holds for $k = 1$ in (20). Suppose that (21) holds for $k \geq 1$. Using (19) and (20) show that (21) holds for $k + 1$:

$$N^{k+1}(p) = N\left(N^k(p)\right) = N(0) + (1 - nN(0))N^k(p) = N(0) + (1 - nN(0))\left(\frac{1}{n} + A^k d\right) =$$
$$N(0) + \frac{1}{n} - nN(0)\frac{1}{n} + (1 - nN(0))A^k d = \frac{1}{n} + A^{k+1}d \ \blacksquare$$

**Theorem 4.** *For linear negator $N$ for any $p$ in [0,1] it holds:*

$$\lim_{k \to \infty} \left(N^k(p)\right) = \frac{1}{n}. \tag{22}$$

**Proof.** From (14) we have $\frac{1}{n} \leq N(0) \leq \frac{1}{n-1}$, hence for $A = 1 - nN(0)$ we obtain: $1 - n\frac{1}{n-1} \leq A \leq 1 - n\frac{1}{n}$, i.e.,

$$-\frac{1}{n-1} \leq A \leq 0, \quad \text{and} \quad |A| < 1. \tag{23}$$

For $d = p - \frac{1}{n}$ and $p \in [0,1]$ we have: $-\frac{1}{n} \leq d \leq \frac{n-1}{n}$. Taking into account these possible values of $A$ and $d$ obtain from (21):

$$\lim_{k \to \infty} \left(N^k(p)\right) = \lim_{k \to \infty}\left(\frac{1}{n} + A^k d\right) = \frac{1}{n} + \lim_{k \to \infty}(A^k d) = \frac{1}{n} + d\lim_{k \to \infty}(A^k) = \frac{1}{n} + d \cdot 0 = \frac{1}{n} \ \blacksquare$$

Definition 1 of linear negation says that the negation $NOT_N(P) = (N(p_1), \ldots, N(p_n))$ of a probability distribution $P = (p_1, \ldots, p_n)$ is a linear negation of pd if $N(p)$ is a linear negator. Define: $NOT_N^k(P) = (N^k(p_1), \ldots, N^k(p_n))$. From the construction of $N^k(p)$ it is clear that $NOT_N^k(P)$ is a probability distribution, i.e., $N^k(p_i)$ satisfies the properties (5). Let us check, for example, that $\sum_{i=1}^n N^k(p_i) = 1$. Indeed, from (21) we have: $\sum_{i=1}^n N^k(p_i) = \sum_{i=1}^n \left(\frac{1}{n} + A^k\left(p_i - \frac{1}{n}\right)\right) = \sum_{i=1}^n \frac{1}{n} + A^k \sum_{i=1}^n \left(p_i - \frac{1}{n}\right) = 1 + A^k \left(\sum_{i=1}^n p_i - \sum_{i=1}^n \frac{1}{n}\right) = 1 + A^k(1-1) = 1.$

From Theorem 4, we have:

**Theorem 5.** *If N is a linear negator, then for the corresponding linear negation of pd* $NOT_N(P) = (N(p_1), \ldots, N(p_n))$ *for any pd P in* $\mathcal{P}_n$ *it holds:*

$$\lim_{k \to \infty} \left( NOT_N^k(P) \right) = NOT_U(P) = P_U.$$

**Proof.** $\lim_{k \to \infty} \left( NOT_N^k(P) \right) = \lim_{k \to \infty} \left( N^k(p_1), \ldots, N^k(p_n) \right) = \left( \lim_{k \to \infty} N^k(p_1), \ldots, \lim_{k \to \infty} N^k(p_n) \right) = \left( \frac{1}{n}, \ldots, \frac{1}{n} \right) = P_U = NOT_U(P)$ ∎

As it follows from Theorem 5, the multiple negations of probability distributions have as the limit the uniform distribution with the maximal entropy value [1]:

$$H(P) = \sum_{i=1}^{n}(1 - p_i)p_i.$$

Results obtained for linear negators also fulfilled for Yager's negator, because it is a linear negator. Hence, for Yager's negator we have from (21): $A = 1 - nN_Y(0) = 1 - n\frac{1-0}{n-1} = -\frac{1}{n-1}$, and for all $k = 0, 1, 2, \ldots$ it is fulfilled:

$$N_Y^k(p) = \frac{1}{n} + (-1)^k \frac{1}{(n-1)^k}\left(p - \frac{1}{n}\right). \tag{24}$$

### 5. Contracting Negators

The concepts of contracting and expanding negations have been introduced and studied in [14, 16-18] on the sets of lexicographic valuations and membership values. Here we extend these concepts on negations of probability values.

**Definition 2.** Let *p* be a probability value from [0,1]. A negator *N* is called *contracting* in *p* if

$$min\{p, N(p)\} \leq N(N(p)) \leq max\{p, N(p)\}, \tag{25}$$

*expanding* in *p* if

$$min\{N(p), N(N(p))\} \leq p \leq max\{N(p), N(N(p))\}, \tag{26}$$

and *involutive* in *p* if

$$N(N(p)) = p. \tag{27}$$

A negator *N* is called *contracting, expanding,* or *involutive* (on [0,1]) if it satisfies the corresponding property for all *p* in [0,1]. A negator *N* is called *non-involutive* if it is not involutive. A negator *N* is called *strictly contracting* if for any $p \neq \frac{1}{n}$ all inequalities in (25) are strict:

$$min\{p, N(p)\} < N(N(p)) < max\{p, N(p)\}. \tag{28}$$

A negation $NOT_N(P) = (N(p_1), \ldots, N(p_n))$ of probability distributions $P = (p_1, \ldots, p_n)$ will be called *contracting* if it is generated by contracting negator *N*.

**Theorem 6.** *Any negator N for any p in [0,1] satisfies (25) or (26), hence it is contracting or expanding in p. N satisfies both properties (25) and (26) if and only if N is involutive in p.*

**Proof.** Suppose $p \leq N(p)$, then from (6), we obtain: $N(N(p)) \leq N(p)$, that gives $p \leq N(N(p)) \leq N(p)$ or $N(N(p)) \leq p \leq N(p)$, and hence (25) or (26), respectively, fulfilled.

Dually, if $N(p) \leq p$, then from (6) we obtain $N(p) \leq N(N(p))$, that gives $N(p) \leq p \leq N(N(p))$ or $N(p) \leq N(N(p)) \leq p$, and hence (26) or (25), respectively, fulfilled.

If $N$ is involutive in $p$ then (25) coincides with (26), and both hold. Suppose both (25) and (26) hold together. If $p \leq N(p)$ then from (6), we obtain $N(N(p)) \leq N(p)$, and from (25) and (26) we obtain $p \leq N(N(p))$ and $N(N(p)) \leq p$ hence (27) is fulfilled. Similarly, we obtain (27) from $N(p) \leq p$, (6), (25) and (26) ∎

**Theorem 7.** Any *pd-independent negator N is non-involutive.*

**Proof.** From (13) and (16) we have: $N(1) \in \left[0, \frac{1}{n}\right]$, and $N(N(1)) \in \left[\frac{1}{n}, \frac{1}{n-1}\right]$, i.e., $N(N(1)) < 1$, hence $N$ is non-involutive ∎

From this theorem, it follows that involutive negators we need to look for between pd-dependent negators. Such negator we introduce in the following section.

A linear negator $N$ such that $N \neq N_U$ will be referred to as a *non-trivial linear negator*.

**Theorem 8.** *Any non-trivial linear negator is strictly contracting.*

**Proof.** From $N \neq N_U$, Corollary 1 and (14) follows $N(p) \neq \frac{1}{n}$ for all $p \neq \frac{1}{n}$, hence $N(0) \neq \frac{1}{n}$, and $\frac{1}{n} < N(0) \leq \frac{1}{n-1}$, and in (21) for $A = 1 - nN(0)$ we have:

$$-\frac{1}{n-1} \leq A < 0, \text{ i.e., A is negative, and } |A| < 1.$$

Using representation (21): $N^k(p) = \frac{1}{n} + A^k d$, and $p = \frac{1}{n} + d$, we need to prove that (28) is satisfied for all $p \neq \frac{1}{n}$ in [0,1], i.e. the following inequalities are fulfilled:

$$min\{p, N(p)\} = min\left\{\frac{1}{n} + d, \frac{1}{n} + Ad\right\} < N(N(p)) = \frac{1}{n} + A^2 d < max\{p, N(p)\} =$$
$$max\left\{\frac{1}{n} + d, \frac{1}{n} + Ad\right\}. \tag{29}$$

If $d > 0$ we have: $\min\left\{\frac{1}{n} + d, \frac{1}{n} + Ad\right\} = \frac{1}{n} + Ad < \frac{1}{n} + A^2 d < \frac{1}{n} + d = \max\left\{\frac{1}{n} + d, \frac{1}{n} + Ad\right\}$, i.e. (29) and hence (28) are satisfied.

If $d < 0$ we have: $\min\left\{\frac{1}{n} + d, \frac{1}{n} + Ad\right\} = \frac{1}{n} + d < \frac{1}{n} + A^2 d < \frac{1}{n} + Ad = \max\left\{\frac{1}{n} + d, \frac{1}{n} + Ad\right\}$, i.e. (29) and hence (28) are satisfied.

Hence linear negator is strictly contracting ∎

Strictly contracting linear negator can be represented by a contracting spiral in Fig. 1, which depicts a sequence of linear negator values $p = N^0(p), N^1(p), N^2(p), N^3(p), \dots$, for $p > \frac{1}{n}$ in the form of a spiral, contracting around the fixed point $\frac{1}{n}$. Fig. 2 depicts this sequence $N^k(p)$ from Theorem 4 with the limit $\frac{1}{n}$.

## 6. Involutive Negators and Involutive Negations

Let $P = (p_1, \dots, p_n)$ be a probability distribution from $\mathcal{P}_n$, and $NOT(P)$ be a negation of pd $P$. It will be called an *involutive* negation if the following property is satisfied:

$$NOT(NOT(P)) = P, \text{ for all } P \text{ in } \mathcal{P}_n. \tag{30}$$

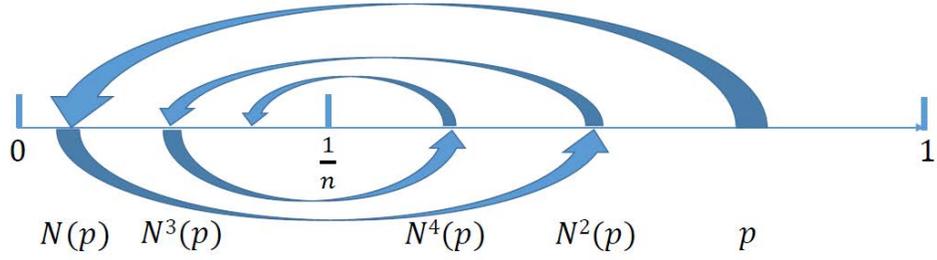

Fig. 1. Contracting negator as a spiral contracting around the fixed point $\frac{1}{n}$

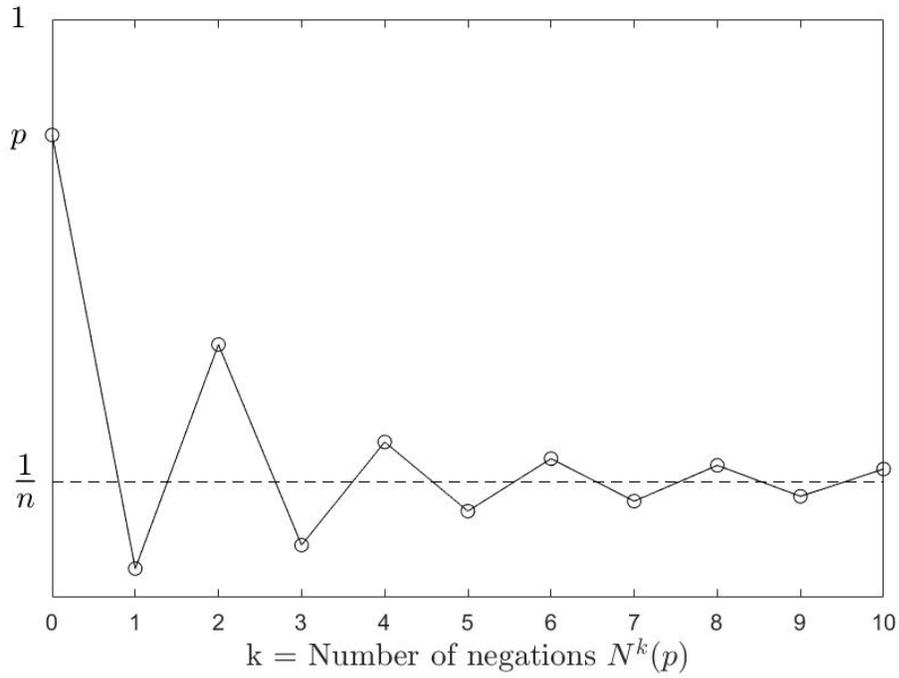

Fig. 2. Contracting negator as a sequence of $N^k(p)$ with the limit $\frac{1}{n}$

A negator $N$ will be called an *involutive* negator if for any pd $P = (p_1, \ldots, p_n)$ it satisfies the following property:

$$N(N(p_i)) = p_i, \text{ for all } i = 1, \ldots, n. \tag{31}$$

It is clear that a negation

$$NOT_N(P) = (N(p_1), \ldots, N(p_n)), \tag{32}$$

will be involutive if it is generated by an involutive negator $N$. Indeed:

$$NOT_N(NOT_N(P)) = NOT_N\big((N(p_1), \ldots, N(p_n))\big) = \big(N(N(p_1)), \ldots, N(N(p_n))\big) = (p_1, \ldots, p_n) = P.$$

All examples of negators considered in the previous section are non-involutive. Here we introduce an involutive negator that will generate the involutive negation of finite discrete probabilistic distributions. As it follows from Theorem 7, an involutive negator cannot be pd-independent.

Let $P = (p_1, \ldots p_n)$ be a probability distribution. Denote $\max(P) = \max_i\{p_i\} = \max\{p_1, \ldots, p_n\}$, $\min(P) = \min_i\{p_i\} = \min\{p_1, \ldots, p_n\}$ and $MP = \max(P) + \min(P)$.

**Theorem 9.** *Let $P = (p_1, \ldots p_n)$ be a probability distribution. Then the function:*

$$N(p_i) = \frac{\max(P)+\min(P)-p_i}{n(\max(P)+\min(P))-1} = \frac{MP-p_i}{nMP-1}, \text{ for all } i = 1, \ldots, n, \tag{33}$$

*is an involutive negator.*

**Proof.** Let us check the fulfillment for (33) the properties (5), (6), and (31).

From (33) and (1) obtain: $\sum_{i=1}^n N(p_i) = \sum_{i=1}^n \frac{MP-p_i}{nMP-1} = \frac{\sum_{i=1}^n MP - \sum_{i=1}^n p_i}{nMP-1} = \frac{nMP-1}{nMP-1} = 1$. From (33) obtain: $N(p_i) \geq 0$, and from $\sum_{i=1}^n N(p_i) = 1$ it follows $N(p_i) \leq 1$ for all $i = 1, \ldots, n$, hence (5) is fulfilled.

From (33), it is clear that $N(p_i)$ is a decreasing function of $p_i$ hence (6) is fulfilled.

Let us prove (31). Denote $Q = NOT_N(P) = \big(N(p_1), \ldots, N(p_n)\big) = (q_1, \ldots, q_n)$, where $q_i = N(p_i)$ for all $i = 1, \ldots, n$. From (33), we have: $q_i = N(p_i) = \frac{MP-p_i}{nMP-1}$, and

$$\max(Q) = \max_i\{N(p_i)\} = \max_i\left\{\frac{MP-p_i}{nMP-1}\right\} = \frac{MP-\min_i\{p_i\}}{nMP-1} = \frac{\max(P)+\min(P)-\min(P)}{nMP-1} = \frac{\max(P)}{nMP-1}.$$

Dually obtain: $\min(Q) = \frac{\min(P)}{nMP-1}$, and:

$$MQ = \max(Q) + \min(Q) = \frac{\max(P)}{nMP-1} + \frac{\min(P)}{nMP-1} = \frac{\max(P)+\min(P)}{nMP-1} = \frac{MP}{nMP-1}. \tag{34}$$

From (33) and (34) obtain:

$$N\big(N(p_i)\big) = N(q_i) = \frac{MQ-q_i}{nMQ-1} = \frac{\frac{MP}{nMP-1}-N(p_i)}{n\frac{MP}{nMP-1}-1} = \frac{\frac{MP}{nMP-1}-\frac{MP-p_i}{nMP-1}}{n\frac{MP}{nMP-1}-1} = \frac{\frac{p_i}{nMP-1}}{\frac{nMP-nMP+1}{nMP-1}} = p_i,$$

hence (31) is fulfilled, and $N$ is involutive. ∎

When $MP = 1$ involutive negator (33) coincides with Yager's negator. This situation appears when $n = 2$ or for a point distribution $P_{(i)} = (p_1, \ldots, p_n)$ satisfying the property: $p_i = 1$ for some $i = 1, \ldots, n$, and $p_j = 0$ for all $j \neq i$. For example, for $P_{(1)} = (1, 0, \ldots, 0)$, we have: $NOT_N(P_{(1)}) = \big(N(1), N(0), \ldots, N(0)\big) = \big(0, \frac{1}{n-1}, \ldots, \frac{1}{n-1}\big)$.

From Proposition 1, it follows that the involutive negator $N$ has a fixed point $p = \frac{1}{n}$, i.e.,

$$N\left(\frac{1}{n}\right) = \frac{1}{n}, \tag{35}$$

and the negation of a probability distribution $NOT_N(P) = \big(N(p_1), \ldots, N(p_n)\big)$ generated by $N$ has a fixed point $P_U = \big(\frac{1}{n}, \ldots, \frac{1}{n}\big)$:

$$NOT_N(P_U) = P_U.$$

The following Proposition says that they are unique fixed points.

**Proposition 6.** *For any pd $P = (p_1, \ldots, p_n)$ the involutive negator $N$ defined by (33) has the unique fixed point $p = \frac{1}{n}$, and the uniform distribution $P_U = \left(\frac{1}{n}, \ldots, \frac{1}{n}\right)$ is the unique fixed point of the involutive negation $NOT_N$ generated by $N$.*

**Proof.** If $p$ is a fixed point of the negator (33) then: $N(p) = \frac{MP - p}{nMP - 1} = p$, that gives sequentially:
$MP - p = p(nMP - 1) = pnMP - p$;  $MP = pnMP$;  $1 = pn$;  and finally: $p = \frac{1}{n}$.

Hence, $p = \frac{1}{n}$ is the unique fixed point of any negator $N$ defined by (33).

Suppose $P = (p_1, \ldots, p_n)$ is a fixed point of $NOT_N$ generated by $N$. Then $NOT_N(P) = (N(p_1), \ldots, N(p_n)) = (p_1, \ldots, p_n)$, hence $N(p_i) = p_i$, for all $i = 1, \ldots, n$, i.e. $p_i$ are fixed points of $N$, hence $p_i = \frac{1}{n}$, and $P = \left(\frac{1}{n}, \ldots, \frac{1}{n}\right) = P_U$. ∎

Since the involutive negation (33) is strictly decreasing function, from $N\left(\frac{1}{n}\right) = \frac{1}{n}$ it follows for any probability distribution $P = (p_1, \ldots, p_n)$ and any $i = 1, \ldots, n$:

$$\begin{cases} if \, p_i < \frac{1}{n} \text{ then } N(p_i) > \frac{1}{n} \\ N\left(\frac{1}{n}\right) = \frac{1}{n} \\ if \, p_i > \frac{1}{n} \text{ then } N(p_i) < \frac{1}{n} \end{cases} \quad (36)$$

If $P = (p_1, \ldots, p_n) \neq \left(\frac{1}{n}, \ldots, \frac{1}{n}\right) = P_U$, i.e., not all $p_i$ equal to $\frac{1}{n}$ then from $\sum_{i=1}^n p_i = 1$ it follows $\min(P) < \frac{1}{n} < \max(P)$, and from (36) and strict monotonicity of $N$ it follows: $N(\min(P)) > \frac{1}{n} > N(\max(P))$ and for all $p_i$, $i = 1, \ldots, n$ it is fulfilled: $\max_i\{N(p_i)\} = N(\min(P)) = \frac{\max(P)}{nMP - 1} \geq N(p_i) \geq N(\max(P)) = \frac{\min(P)}{nMP - 1} = \min_i\{N(p_i)\}$.

To summarize the last considerations, we can say that for any probability distribution $P = (p_1, \ldots, p_n)$ the values of negations $N(p_i)$, $i = 1, \ldots, n$, are located on the line connecting two points in 2-dimentional space with coordinates: $\left(\min(P), \frac{\max(P)}{nMP - 1}\right)$ and $\left(\max(P), \frac{\min(P)}{nMP - 1}\right)$.

**Example 1**. Consider probability distribution $P = (0.1, 0.2, 0.15, 0.3, 0.25)$. We have $n = 5$, fixed point: $\frac{1}{n} = \frac{1}{5} = 0.2$, and $N(0.2) = 0.2$, $MaxP = 0.3$, $MinP = 0.1$, $MP = 0.4$, $nMP - 1 = 5(0.4) - 1 = 1$, $N(p_i) = \frac{MP - p_i}{nMP - 1} = \frac{0.4 - p_i}{1} = 0.4 - p_i$.  $NOT_N(P) = (N(0.1), N(0.2), N(0.15), N(0.3), N(0.25)) = (0.3, 0.2, 0.25, 0.1, 0.15)$. After similar calculations, we obtain: $NOT_N(NOT_N(P)) = P$.

## 7. Conclusion

The paper studied negators generating element-by-element negations of probability distributions (pd). We showed that the sequence of multiple negations of pd generated by a pd-independent linear negator converges to the uniform distribution with maximal entropy. We showed that all pd-independent negators are non-involutive, and non-trivial linear negators are strictly contracting; hence, we need to look for an involutive negator in the class of pd-dependent negators. Finally, we introduced an involutive negator in the class of pd-dependent negators. It generates involutive negation of probability distributions. Such involutive negation can formalize a probability distribution NOT P, where P is some linguistic concept like HIGH defined on a set of probability distributions. The involutivity of negation, like NOT(NOT(P)) = P, is a common property for many logical systems and can be used in retrieval or reasoning systems operating with terms represented by probability distributions. We plan to apply the proposed involutive negation in Dempster-Shaffer theory as it was proposed in the original work of Yager [1].

**Acknowledgments.** The work is partially supported by SIP project 20211874 of IPN, Mexico.